\documentclass[11pt]{article}
\pdfoutput=1
\usepackage{CJKutf8}

\usepackage{times}
\usepackage{url}
\usepackage{color}
\usepackage{float}
\usepackage{tabularx}
\usepackage{subfig}
\usepackage{multirow}
\usepackage{latexsym}
\usepackage{amsmath}
\usepackage{amssymb}
\usepackage{multirow}
\usepackage[lined,boxed,commentsnumbered]{algorithm2e}

\usepackage{pgfplots}
\title{Overview of the NLPCC 2015 Shared Task:\\
Chinese Word Segmentation and POS Tagging for Micro-blog Texts
}
\author{Xipeng Qiu, Peng Qian, Liusong Yin, Shiyu Wu, Xuanjing Huang\\
School of Computer Science, Fudan University\\
825 Zhangheng Road, Shanghai, China\\
\{xpqiu,pqian11,lsyin14,sywu13,xjhuang\}@fudan.edu.cn}

\date{}

\begin{document}

\maketitle

\begin{abstract}

In this paper, we give an overview for the shared task at the 4th CCF Conference on Natural Language Processing \& Chinese Computing (NLPCC 2015):
Chinese word segmentation and part-of-speech (POS) tagging for micro-blog texts.
Different with the popular used newswire datasets, the dataset of this shared task consists of the relatively informal micro-texts. The shared task has two sub-tasks: (1) individual Chinese word segmentation and (2) joint Chinese word segmentation and POS Tagging. Each subtask has three tracks to distinguish the systems with different resources. We first introduce the dataset and task, then we characterize the different approaches of the participating systems, report the test results, and provide a overview analysis of these results.
An online system is available for open registration and evaluation at \url{http://nlp.fudan.edu.cn/nlpcc2015}.
\end{abstract}

\section{Introduction}

Word segmentation and Part-of-Speech (POS) tagging are two fundamental tasks for Chinese
language processing. In recent years, word segmentation and POS tagging have undergone great
development. The popular method is to regard these two tasks as sequence labeling problem \cite{Xue:2003,peng:2004},
which can be handled with supervised learning algorithms such as Maximum Entropy (ME) \cite{Berger:1996}, averaged perceptron \cite{Collins:2002}, Conditional Random Fields
(CRF)\cite{Lafferty:2001}. After years of intensive researches, Chinese word segmentation and POS tagging achieve a quite high precision.  However, their performance is not so satisfying for the practical demands to analyze Chinese texts, especially for informal texts. The key reason is that most of annotated corpora are drawn from news texts. Therefore, the system trained on these corpora cannot work well with the out-of-domain texts.

In this shared task, we focus to evaluate the performances of word segmentation and POS tagging on relatively informal micro-texts.

\section{Data}
Different with the popular used newswire dataset, we use relatively informal texts from Sina
Weibo\footnote{http://weibo.com/}. The training and test data consist of micro-blogs from
various topics, such as finance, sports, entertainment, and so on.  Both the training and test files are UTF-8 encoded.

The information of dataset is shown in Table \ref{tb:dataset}. The out-of-vocabulary (OOV) rate is slight higher than the other benchmark datasets. For example, the OOV rate is $5.58\%$ in the popular division \cite{Yang:2012} of the Chinese Treebank (CTB 6.0) dataset \cite{Xue:2005a}, while the OOV rate of our dataset is $7.25\%$.

\begin{CJK*}{UTF8}{gbsn}  
\begin{table}[H]
  \centering
\caption{Statistical information of dataset.}\label{tb:dataset}
  \begin{tabular}{|c|c|c|c|c|c|c|}
\hline
Dataset & Sents & Words & Chars & Word Types & Char Types & OOV Rate \\\hline
Training  & 10,000 & 215,027 & 347,984 & 28,208 & 39,71 & - \\\hline
Test & 5,000 & 106,327 & 171,652 & 18,696 & 3,538 & 7.25\% \\\hline
Total & 15,000 & 322,410 & 520,555 & 35,277 & 4,243 & -\\\hline
\end{tabular}
\end{table}

There are total 35 POS tags in this dataset. A detailed list of POS tags is shown in Table \ref{tb:pos}.

\begin{table}[H]\footnotesize
  \centering
  \caption{Statistical information of POS tags.}\label{tb:pos}

  \begin{minipage}{.45\linewidth}
  \begin{tabular}{|l|l|l|r|}
\hline
\multicolumn{2}{|c|}{词性(POS)}		&	En & Num \\\hline

\multicolumn{2}{|c|}{名词} &		NN	&84,006\\\hline
\multirow{8}{*}{实体名}
&	人名	&PER&	3,232\\\cline{2-4}
&	机构名	&ORG&	2,578\\\cline{2-4}
&	地名	&LOC&	9,701\\\cline{2-4}
&	其他	&NR	&  550\\\cline{2-4}
&	邮件	&EML&	3\\\cline{2-4}
&	型号名	&MOD&	34\\\cline{2-4}
%&	专有名	&SN	&92\\\cline{2-4}
&	网址	&URL&	11\\\hline

\multirow{2}{*}{副词}
&	疑问副词 &	ADQ&	340\\\cline{2-4}
&	副词	& AD &	26,155 \\\hline
\multirow{2}{*}{形貌}
&	形容词	&JJ&	9,477\\\cline{2-4}
&	形谓词	&VA	& 3,339\\\hline
\multirow{5}{*}{动词}
&	动词&	VV	&51,294\\\cline{2-4}
&	情态词&	MV&	3,700 \\\cline{2-4}
&	趋向动词&	DV&	781 \\\cline{2-4}
&	被动词&	BEI&	927\\\cline{2-4}
&	把动词&	BA&	600\\\hline
\multicolumn{2}{|c|}{时间短语}	&	NT&	5,881\\\hline
\end{tabular}
\end{minipage}\begin{minipage}{.45\linewidth}
\begin{tabular}{|l|l|l|r|}
\hline
\multicolumn{2}{|c|}{词性(POS)}		&	Labels & Num \\\hline
\multirow{3}{*}{代词}
&	人称代词	&PNP	&4,903\\\cline{2-4}
&	疑问代词	&PNQ	&492\\\cline{2-4}
&	指示代词	&PNI	&834\\\cline{2-4}
\multirow{2}{*}{连词}
&	并列连词&	CC&	2,725\\\cline{2-4}
&	从属连词&	CS	& 866\\\hline
\multirow{3}{*}{数量}
&	数词	&CD&	10,764\\\cline{2-4}
&	量词	&M&	7,917\\\cline{2-4}
&	序数词	&OD&	1,219\\\hline

\multirow{9}{*}{助词}
&	方位词	&LC	&4,725\\\cline{2-4}
&	省略词&	ETC	&673\\\cline{2-4}
&	语气词	&SP	&1,076\\\cline{2-4}
&	限定词&	DT	&3,579\\\cline{2-4}
&	叹词&	IJ	&20\\\cline{2-4}
&	标点&	PU	&52,922 \\\cline{2-4}
&	结构助词&	DSP	&13,756\\\cline{2-4}
&	介词&	P	&9,488\\\cline{2-4}
&	时态词&	AS&	3,382\\\hline
\end{tabular}
\end{minipage}
\end{table}
\end{CJK*}

\subsection{Background Data}
Besides the training data, we also provide the background data, from which the training and
test data are drawn. The purpose is to find the more sophisticated features by the
unsupervised way.

\section{Description of the Task}

 In this shared task, we wish to investigate
the performances of Chinese word segmentation and POS tagging for the micro-blog texts.

\subsection{Subtasks}
This task focus the two fundamental problems of Chinese language processing: word
segmentation and POS tagging, which can be divided into two subtasks:
\begin{enumerate}
  \item \textbf{SEG} Chinese word segmentation
  \item \textbf{S\&T} Joint Chinese word segmentation and POS Tagging
\end{enumerate}

\subsection{Tracks}

Each participant will be allowed to submit the three runs for each subtask: \textbf{closed
track} run, \textbf{semi-open track}  run and \textbf{open track}  run.
\begin{enumerate}
  \item In the \textbf{closed} track, participants could only
use information found in the provided training data. Information such as externally
obtained word counts, part of speech information, or name lists was excluded.
  \item In the \textbf{semi-open} track, participants could use the information extracted from the
      provided background data in addition to the provided training data. Information such
      as externally obtained word counts, part of speech information, or name lists was
      excluded.
    \item In the \textbf{open} track, participants could use the information which should be public
        and be easily obtained. But it is not allowed to obtain the result by the manual
        labeling or crowdsourcing way.
\end{enumerate}

\section{Participants}

Sixteen teams have registered for this task. Finally, there are 27 qualified submitted results from 10 teams. A summary of qualified participating teams are shown in Table \ref{tb:team}.

\begin{table}[H]
  \centering
\caption{Summary of the participants.}\label{tb:team}
\begin{tabular}{|*{7}{c|}}
  \hline
  & \multicolumn{3}{|c|}{SEG}	 & 	 \multicolumn{3}{|c|}{S\&T} \\\hline
 & 	closed & 	open & 	semi-open & 	closed & 	open & 	semi-open \\\hline
NJU & 	$\surd$ & 	$\surd$ & 	$\surd$ & 	 & 	 & 	 \\\hline
BosonNLP & 	$\surd$ & 	$\surd$ & 	 & 	$\surd$ & 	$\surd$ & 	 \\\hline
CIST & 	$\surd$ & 	 & 	$\surd$ & 	$\surd$ & 	 & 	$\surd$ \\\hline
XUPT & 	$\surd$ & 	 & 	 & 	$\surd$ & 	 & 	 \\\hline
CCNU & 	$\surd$ & 	$\surd$ & 	 & 	 & 	 & 	 \\\hline
ICT-NLP & 	$\surd$ & 	 & 	 & 	 & 	 & 	 \\\hline
BJTU & 	$\surd$ & 	$\surd$ & 	$\surd$ & 	$\surd$ & 	$\surd$ & 	$\surd$ \\\hline
SZU & 	 & 	$\surd$ & 	 & 	 & 	$\surd$ & 	 \\\hline
ZZU & 	 & 	 & 	$\surd$ & 	 & 	 & 	 \\\hline
WHU & 	 & 	 & 	 & 	$\surd$ & 	 & 	$\surd$ \\\hline
\end{tabular}
\end{table}

\section{Results}

\subsection{Evaluation Metrics}
The evaluation measure are reported are precision, recall, and an evenly-weighted F1.

\subsection{Baseline Systems}

Currently, the mainstream method of word segmentation is discriminative character-based sequence labeling. Each character is labeled as one of \{B, M, E, S\} to indicate the segmentation. \{B, M, E\} represent \textit{Begin}, \textit{Middle}, \textit{End} of a multi-character segmentation respectively, and S represents a \textit{Single} character segmentation. 

For the joint word segmentation and POS tagging, the state-of-the-art method is also based on sequence learning with cross-labels, which can avoid the problem of error propagation and achieve higher performance on both subtasks\cite{Ng:2004}. Each label is the cross-product of a segmentation label and a tagging label, e.g. \{B-NN, I-NN, E-NN, S-NN, ...\}. The features are generated by position-based templates on character-level.

Sequence labeling is the task of assigning labels $\mathbf{y} = y_1, \dots, y_n$  to an input sequence $\mathbf{x} = x_1, \dots ,x_n$. Given a sample $\mathbf{x}$, we define the feature $\Phi(\mathbf{x},\mathbf{y})$. Thus, we can label $\mathbf{x}$ with a score function,
\begin{equation}\label{}
 \hat{\mathbf{y}} = \arg\max_\mathbf{y} F(\mathbf{w},\Phi(\mathbf{x},\mathbf{y})),
\end{equation}
where $\mathbf{w}$ is the parameter of function $F(\cdot)$.

For sequence labeling, the feature can be denoted as $\phi_k({y_i},{y_{i-1}},\mathbf{x},{i})$, where ${i}$ stands for the position in the sequence and ${k}$ stands for the number of feature templates.

Here, we use two popular open source toolkits for sequence labeling task as the baseline systems: FNLP\footnote{\url{https://github.com/xpqiu/fnlp/}} \cite{Qiu:2013} and CRF++\footnote{\url{http://taku910.github.io/crfpp/}}.
Here, we use the default setting of CRF++ toolkit with the feature templates as shown in Table \ref{tab:template_crf}. The same feature templates are also used for FNLP.

\begin{table}
\centering
\caption{Templates of CRF++ and FNLP.}\label{tab:template_crf}
\begin{tabular}{|l|p{0.25\textwidth}|}
\hline
unigram feature&$c_{-2}$, $c_{-1}$, $c_{0}$, $c_{+1}$, $c_{+2}$\\
\hline
bigram feature&$c_{-1}\circ c_{0}$, $c_{0}\circ c_{+1}$\\
\hline
trigram feature&$c_{-2} \circ c_{-1}\circ c_{0}$, $c_{-1} \circ c_{0}\circ c_{+1}$, $c_{0}\circ c_{+1} \circ c_{+2}$\\
\hline
\end{tabular}
\end{table}

\subsection{Chinese word segmentation}

In word segmentation task, the best F1 performances are $95.12$, $95.52$ and $96.65$ for closed, semi-open and open tracks respectively. The best system outperforms the baseline systems on closed track. The best system on semi-open track is better than that on closed track. Unsurprisingly, the performances boost greatly on open track.

\begin{table}[H]
  \centering
  \caption{Performances of word segmentation.}\label{tb:seg}
\begin{tabular}{|*{5}{c|}}
  \hline
Systems & 	Precision & 	Recall & 	F1 & 	Track \\\hline
CRF++	&93.3	&93.2	&93.3 & 	\multirow{2}*{baseline, closed}	\\\cline{1-4}
FNLP	&94.1	&93.9	&94.0	& 	 \\\hline\hline
 
NJU & 	95.14 & 	95.09 & 	95.12 & \multirow{7}*{closed}	\\\cline{1-4}
BosonNLP & 	95.03 & 	95.03 & 	95.03 & 	 \\\cline{1-4}
CIST & 	94.78 & 	94.42 & 	94.6 &	 \\\cline{1-4}
XUPT & 	94.61 & 	93.85 & 	94.22 &	 \\\cline{1-4}
CCNU & 	93.95 & 	93.45 & 	93.7 &	 \\\cline{1-4}
ICT-NLP & 	93.96 & 	92.91 & 	93.43 & 	 \\\cline{1-4}
BJTU & 	89.49 & 	93.55 & 	91.48 & 	 \\\hline\hline
CIST & 	95.47 & 	95.57 & 	95.52 & 	\multirow{4}*{ semi-open}		 \\\cline{1-4} 
NJU & 	95.3 & 	95.31 & 	95.3 & 		 \\\cline{1-4}
BJTU & 	90.91 & 	94.46 & 	92.65 & 	 \\\cline{1-4}
ZZU & 	85.36 & 	85.25 & 	85.31 &  \\\hline\hline
BosonNLP & 	96.56 & 	96.75 & 	96.65 & \multirow{5}*{ open}		 \\\cline{1-4}
NJU & 	96.03 & 	96.15 & 	96.09 & 	 \\\cline{1-4}
SZU & 	95.52 & 	95.64 & 	95.58 & 	 \\\cline{1-4}
CCNU & 	93.68 & 	93.09 & 	93.38 & 	 \\\cline{1-4}
BJTU & 	91.79 & 	94.92 & 	93.33 & 		 \\\hline
\end{tabular}

\end{table}

\subsection{Joint Chinese word segmentation and POS Tagging}

In the joint word segmentation and POS tagging, the best performances are $88.93$, $88.69$ and $91.55$ for closed, semi-open and open tracks respectively.

\begin{table}[H]
  \centering
  \caption{Performances of joint word segmentation and POS tagging.}\label{tb:seg}

\begin{tabular}{|*{5}{c|}}
  \hline
Systems & 	Precision & 	Recall & 	F1 & 	Track \\\hline
%CRF++	&93.3	&93.2	&93.3 & 	\multirow{2}*{baseline, closed}	\\\cline{1-4}
%FNLP	&94.1	&93.9	&94.0	& 	 \\\hline\hline
BosonNLP & 	88.91 & 	88.95 & 	88.93 & \multirow{5}*{closed}	\\\cline{1-4}
XUPT & 	88.54 & 	87.83 & 	88.19 & 	\\\cline{1-4}
BJTU & 	88.28 & 	87.67 & 	87.97 & 	\\\cline{1-4}
CIST & 	88.09 & 	87.76 & 	87.92 & 	\\\cline{1-4}
BJTU & 	80.64 & 	85.1 & 	82.81 & 	 \\\hline\hline
CIST & 	88.64 & 	88.73 & 	88.69 & 	\multirow{3}*{semi-open} \\\cline{1-4}
WHU & 	88.59 & 	87.96 & 	88.27 & 	\\\cline{1-4}
BJTU & 	81.76 & 	85.82 & 	83.74 & 	\\\hline\hline
BosonNLP & 	91.42 & 	91.68 & 	91.55 & 	\multirow{3}*{open} \\\cline{1-4}
SZU & 	88.93 & 	89.05 & 	88.99 & 	 \\\cline{1-4}
BJTU & 	79.85 & 	83.51 & 	81.64 & 	 \\\hline

\end{tabular}
\end{table}

\section{Analysis}

\section{Conclusion}

After years of intensive researches, Chinese word segmentation and POS tagging have achieved a quite high precision.  However, the performances of the state-of-the-art systems are still relatively low for the informal texts, such as micro-blogs, forums. The NLPCC 2015 Shared Task on Chinese Word Segmentation and POS Tagging for Micro-blog Texts
focuses on the fundamental research in Chinese language processing.

It is the first time to use the micro-texts to evaluate the performance of the state-of-the-art methods

In future work, we hope to run an online evaluation system to accept open registration and submission. Currently, a simple system is  available at \url{http://nlp.fudan.edu.cn/nlpcc2015}. The system also gives the leaderboards for the up-to-date results  under the different tasks and tracks.
Besides, we also wish to extend the scale of corpus and add more informal texts.

%There is clearly no single best system. And the participating sites S1, S10, S9, S6, S5 and S18 have all achieved respectable scores on different track runs of this bakeoff. An improvement on the OOV recall over the prior bakeoffs has been observed. It is the first time to apply

\section*{Acknowledgement}
We are very grateful to the students from our lab for their efforts to annotate and check the data. We would also like to thank the participants for their valuable feedbacks and comments.

\bibliographystyle{plain}

\bibliography{../nlp,../ours}

\begin{thebibliography}{1}

\bibitem{Berger:1996}
A.L. Berger, V.J. Della~Pietra, and S.A. Della~Pietra.
\newblock A maximum entropy approach to natural language processing.
\newblock {\em Computational Linguistics}, 22(1):39--71, 1996.

\bibitem{Collins:2002}
Michael Collins.
\newblock Discriminative training methods for hidden markov models: Theory and
  experiments with perceptron algorithms.
\newblock In {\em Proceedings of the 2002 Conference on Empirical Methods in
  Natural Language Processing}, 2002.

\bibitem{Lafferty:2001}
John~D. Lafferty, Andrew McCallum, and Fernando C.~N. Pereira.
\newblock Conditional random fields: Probabilistic models for segmenting and
  labeling sequence data.
\newblock In {\em Proceedings of the Eighteenth International Conference on
  Machine Learning}, 2001.

\bibitem{Ng:2004}
H.T. Ng and J.K. Low.
\newblock Chinese part-of-speech tagging: one-at-a-time or all-at-once?
  word-based or character-based.
\newblock In {\em Proceedings of EMNLP}, volume~4, 2004.

\bibitem{peng:2004}
F.~Peng, F.~Feng, and A.~McCallum.
\newblock Chinese segmentation and new word detection using conditional random
  fields.
\newblock {\em Proceedings of the 20th international conference on
  Computational Linguistics}, 2004.

\bibitem{Qiu:2013}
Xipeng Qiu, Qi~Zhang, and Xuanjing Huang.
\newblock {FudanNLP}: A toolkit for {Chinese} natural language processing.
\newblock In {\em Proceedings of Annual Meeting of the Association for
  Computational Linguistics}, 2013.

\bibitem{Xue:2003}
N.~Xue.
\newblock Chinese word segmentation as character tagging.
\newblock {\em Computational Linguistics and Chinese Language Processing},
  8(1):29--48, 2003.

\bibitem{Xue:2005a}
Naiwen Xue, Fei Xia, Fu-Dong Chiou, and Martha Palmer.
\newblock The {Penn Chinese TreeBank}: Phrase structure annotation of a large
  corpus.
\newblock {\em Natural language engineering}, 11(2):207--238, 2005.

\bibitem{Yang:2012}
Yaqin Yang and Nianwen Xue.
\newblock Chinese comma disambiguation for discourse analysis.
\newblock In {\em Proceedings of the 50th Annual Meeting of the Association for
  Computational Linguistics: Long Papers-Volume 1}, pages 786--794. Association
  for Computational Linguistics, 2012.

\end{thebibliography}
\end{document}